\documentclass[a4paper,twoside]{article}

\usepackage{epsfig}
\usepackage{subcaption}
\usepackage{calc}
\usepackage{amssymb}
\usepackage{amstext}
\usepackage{amsmath}
\usepackage{amsthm}
\usepackage{multicol}
\usepackage{pslatex}
\usepackage{pslatex}
\usepackage{apalike}

\usepackage{multirow}
\usepackage{xcolor}
\usepackage{cite}
\usepackage{enumerate}
\usepackage{url}
\usepackage{balance} % menotti on 10/16/2019
\usepackage{bm}

\usepackage{SCITEPRESS}

\begin{document}

\title{CNN Hyperparameter tuning applied to Iris Liveness Detection}

\author{\authorname{
Gabriela Yukari Kimura\sup{1}\orcidAuthor{0000-0000-0000-0000},
Diego Rafael Lucio\sup{1}\orcidAuthor{0000-0003-2012-3676},
Alceu S. Britto Jr.\sup{2}\orcidAuthor{0000-0002-3064-3563},
and David Menotti\sup{1}\orcidAuthor{0000-0003-2430-2030}}
\affiliation{\sup{1}Laboratory of Vision, Robotics and Imaging, Federal University of Paran\'a, Curitiba, Brazil}
\affiliation{\sup{2}Postgraduate Program in Informatics, Pontifical Catholic University of Paran\'a, Curitiba, Brazil} \vspace{1mm}
\email{
\sup{1}gykimura10@gmail.com, \sup{1}drlucio@inf.ufpr.br, \sup{2}alceu@ppgia.pucpr.br, \sup{1}menotti@inf.ufpr.br}
}

\keywords{Iris Biometrics; Deep Learning; Convolutional Networks}

\abstract{
The iris pattern has significantly improved the biometric recognition field due to its high level of stability and uniqueness. Such physical feature has played an important role in security and other related areas. 
However, presentation attacks, also known as spoofing techniques, can be used to bypass the biometric system with artifacts such as printed images, artificial eyes, and textured contact lenses.
To improve the security of these systems, many liveness detection methods have been proposed, and the first Internacional Iris Liveness Detection competition was launched in 2013 to evaluate their effectiveness. 
In this paper, we propose a hyperparameter tuning of the CASIA algorithm, submitted by the Chinese Academy of Sciences to the third competition of Iris Liveness Detection, in 2017. The modifications proposed promoted an overall improvement, with 8.48\% Attack Presentation Classification Error Rate (APCER) and 0.18\% Bonafide Presentation Classification Error Rate (BPCER) for the evaluation of the combined datasets. Other threshold values were evaluated in an attempt to reduce the trade-off between the APCER and the BPCER on the evaluated datasets and worked out successfully.
}

\onecolumn \maketitle \normalsize \setcounter{footnote}{0} \vfill

%%%%%%%%%%%%%%%%%%%%%%%%%%%%%%%%%%%%%%%%%%%%%%%%%%%%%%%%%%%%%%%%%%%%%%%%%%%%%%%%
\section{\uppercase{INTRODUCTION}}

Biometric recognition offers a natural and reliable alternative for the automatic identification of individuals based on their physiological or behavioral characteristics (e.g. fingerprint, iris, gait, voice, hand geometry, etc.) \cite{jain:2011}.
The iris pattern is regarded as one of the most accurate biometrics owing to its high level of stability and uniqueness. It has played an important role in security and other associated fields \cite{tisse:2002}.

\begin{figure}[!ht]
    \centering
    \begin{tabular}{@{}cc@{}}
    \includegraphics[width=.47\linewidth]{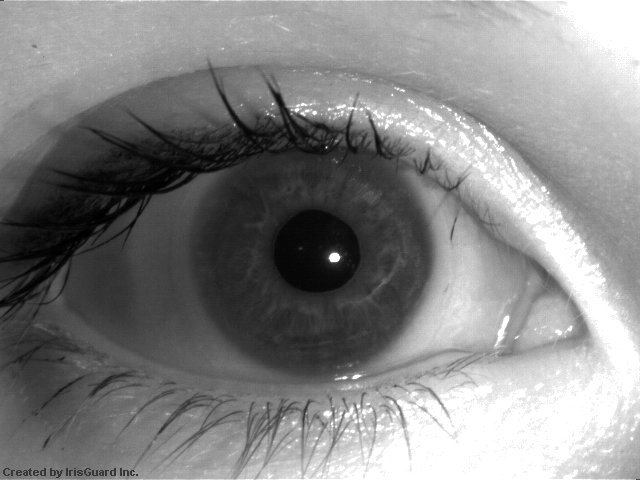} &
    \includegraphics[width=.47\linewidth]{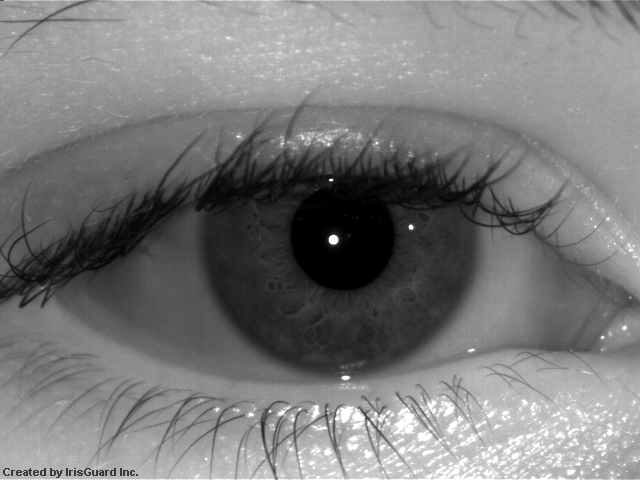} \\
    \includegraphics[width=.47\linewidth]{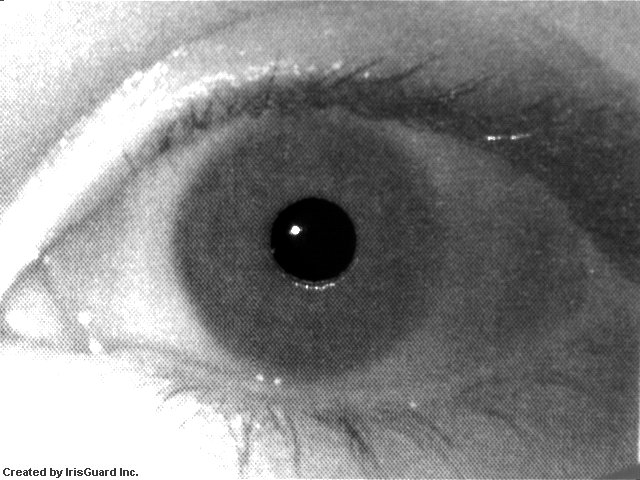} &
    \includegraphics[width=.47\linewidth]{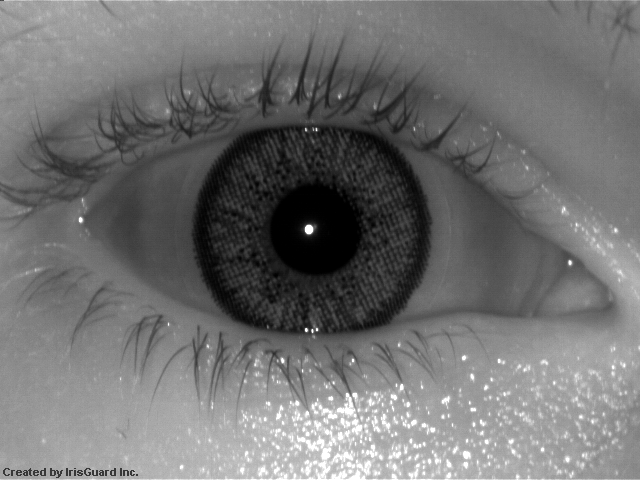} \\
    \includegraphics[width=.47\linewidth]{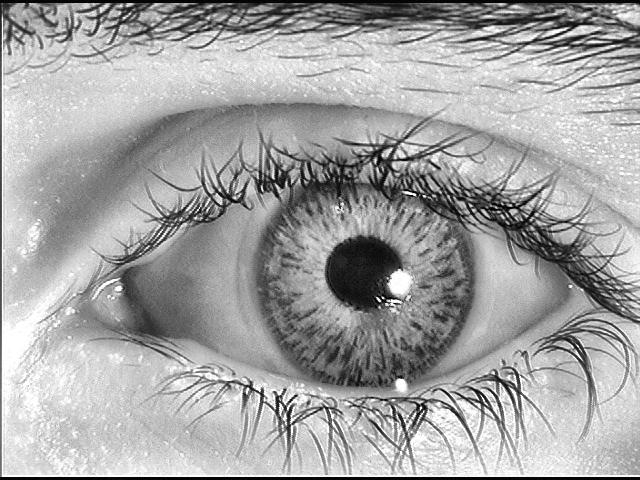} &    
    \includegraphics[width=.47\linewidth]{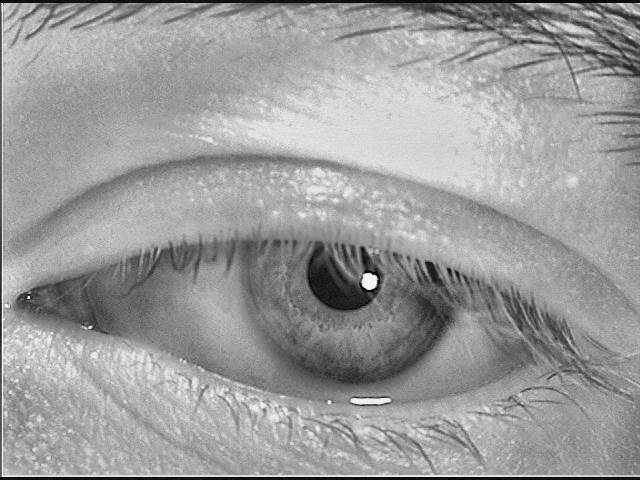} \\    
    \end{tabular}
    \caption{Spoofing problem. Which images are true or fake? Sorting from left to right and from top to bottom, the ones numbered as 1, 4 and 6 are the real ones.}
    \label{fig:spoofingproblem}
\end{figure}

However, despite the many advantages, iris biometric systems are highly susceptible to presentation attacks, usually referred to as spoofing techniques, that attempt to conceal or impersonate other identities \cite{kohli:2016,menotti:2015,toosi:2017,Pala:2017,Czajka:2018,sajjad:2019,tolosana:2020}. 
Examples of typical iris spoofing attacks include printed iris images, video playbacks, artificial eyes, and textured contact lenses. 
Figure~\ref{fig:spoofingproblem} presents some examples.
We observe how difficult is to a human being make the right judgment.

Therefore, in order to enhance the security of iris recognition systems, liveness detection methods have been suggested and can be categorized into software-based and hardware-based techniques. In software-based techniques, the spoof attacks are detected on the sample that has already been acquired with a standard sensor. Besides, they have the advantage of being less expensive and less intrusive to the user. 

In hardware-based techniques, physical characteristics of the iris are measured, such as the behavior of the pupil under different light conditions. Although this approach usually presents a higher detection rate, it has the disadvantage of being expensive and requiring intervention at the device level \cite{r4}.

To evaluate the efficiency of liveness detection algorithms, the first international iris liveness contest was introduced in 2013, by Clarkson University, Notre Dame University and Warsaw University of Technology. 
The third contest was launched in 2017, and included the combined dataset from West Virginia University and IIIT-Delhi\footnote{But we could not reach the authors for obtaining this last dataset}.

The competition was open to all industrial and academic institutions, and three software-based algorithms were submitted for evaluation. They were evaluated by four datasets of live and spoof iris images, represented by printed iris images, patterned contact lenses, and printouts of patterned contact lenses. The algorithm CASIA,  from the Chinese Academy of Sciences, designed a Cascade SpoofNets for iris liveness detection and obtained the second-best results with a combined error rate of 10.68\% with an APCER of 11.88\% and 9.47\% BPCER, while the winner team, an Anonymous submission, presented an APCER of 14.71\% and BPCER of 3.36\% \cite{livdev:2017}.

In this paper, a parameter tuning of the CASIA SpoofNet is proposed. 
Through experiments, the optimal values of batch size, number of epochs, loss function, learning rate, and weight decay were selected. 
The suggested changes achieved better results overall, with the exception of the Notre Dame dataset evaluation, where the modified algorithm presented a higher APCER.

The rest of the paper is structured as follows. 
In Section II, other iris spoofing techniques are reviewed. 
In Section III, the databases used in our experiments are described. 
In Section IV, the proposed methodology is detailed. 
Experimental results are described and discussed in Section V.
Conclusions are finally drawn in Section VI.

\section{\uppercase{Related Work}}

Over the past few years, the vulnerability of iris authentication systems to spoofing techniques has been highly researched. In 2016, Kohli et al.\cite{kohli:2016} proposed a spoofing detection technique based on the textural analysis of the iris images. The method presented a unified framework with feature-level fusion and combines the multi-order Zernike moments, that encodes variations in the structure of the image, with the Local Binary Pattern Variance (LBPV), used for representing textural changes. 
In \cite{r7}, the authors introduced three solutions on printed eye images based on the analysis of image frequency spectrum, controlled light reflection from the cornea, and the behavior of the iris pupil under light conditions. 
And in \cite{r8}, one may find a method using collimated IR-LED (Infra-Red Light Emitting Diode), where the theoretical positions and distances between the Purkinje images are calculated based on the human eye model. 
Both techniques depend on hardware device resolution and in \cite{r7}, the subject cooperation is also required.  

Recent works include Silva et al.~\cite{r9}, which introduced a deep learning technique. 
The area has been evolving and showing promising results in several computer vision problems, such as pedestrian detection\cite{r13}, character recognition \cite{laroca:2018,laroca:2019,laroca:2019b} and face recognition\cite{r14}.

% - A Robust Real-Time Automatic License Plate Recognition Based on the YOLO Detector (DOI: 10.1109/IJCNN.2018.8489629)
% - Convolutional Neural Networks for Automatic Meter Reading (DOI: 10.1117/1.JEI.28.1.013023)
% - An Efficient and Layout-Independent Automatic License Plate Recognition System Based on the YOLO Detector (arXiv)

Their approach addresses three-class image detection problems (textured contact lenses, soft contact lenses, and no lenses), and uses a convolutional network in order to build a deep image representation and an additional fully-connected single layer with softmax regression for classification.

\section{\uppercase{Datasets}}

In this section, the databases used in the experiments are described. All of them are publicly available upon request and were used for the Iris Liveness Detection competition of 2017. 
The images of the databases are in grayscale and their dimensions are 640$\times$480 pixels.
Additional details are presented in the following subsections.

\subsection{Clarkson Dataset}

The Clarkson dataset consists of three sections of images captured by an LG IrisAccess EOU2200 camera. The first and second are live and patterned contacts iris images and the third is composed of printouts of live NIR iris images as well as printouts created from visible light images of the eye captured with an iPhone 5.

\begin{figure}[!ht]
    \centering
    \begin{tabular}{@{}cc@{}}
    \includegraphics[width=.47\linewidth]{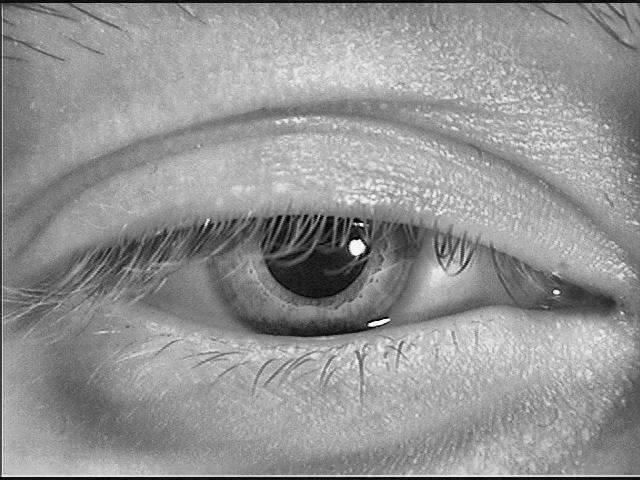} &
    \includegraphics[width=.47\linewidth]{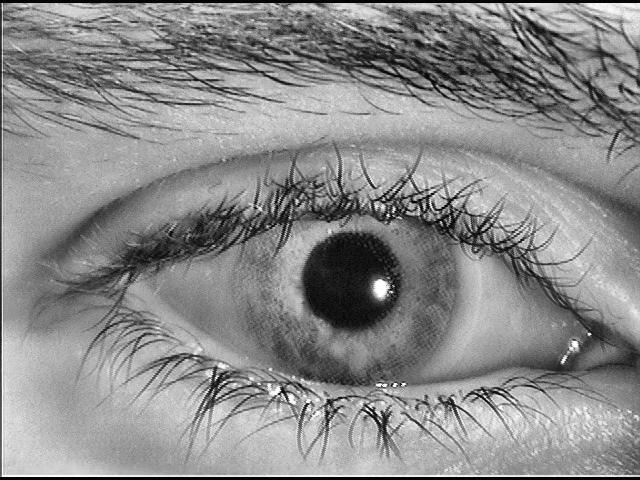} \\
    \includegraphics[width=.47\linewidth]{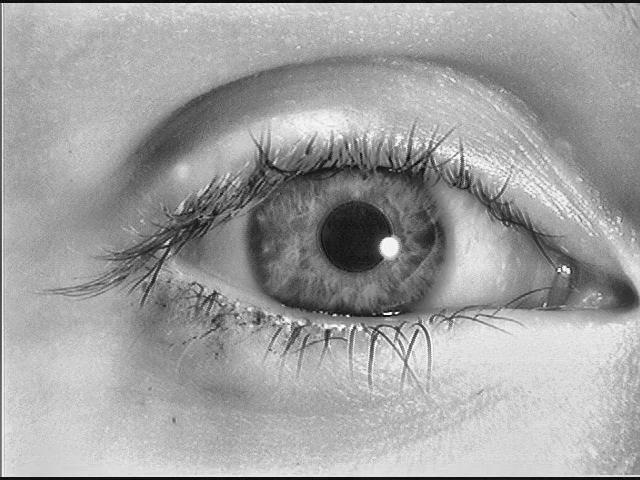} &
    \includegraphics[width=.47\linewidth]{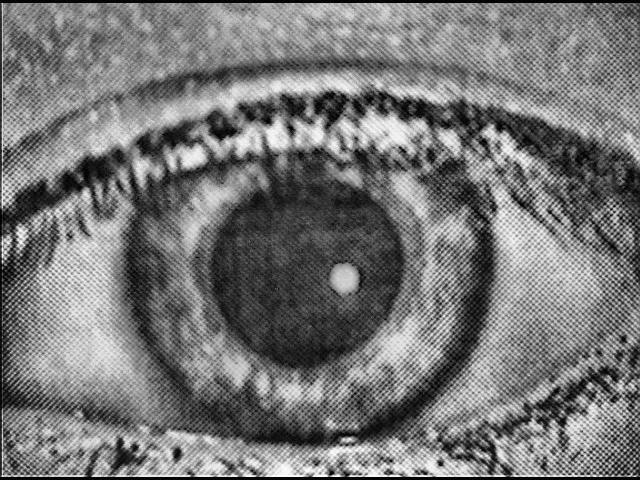} \\
    \includegraphics[width=.47\linewidth]{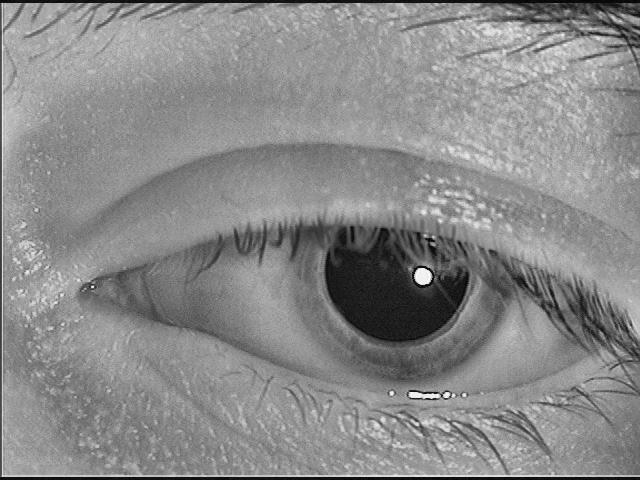} &
    \includegraphics[width=.47\linewidth]{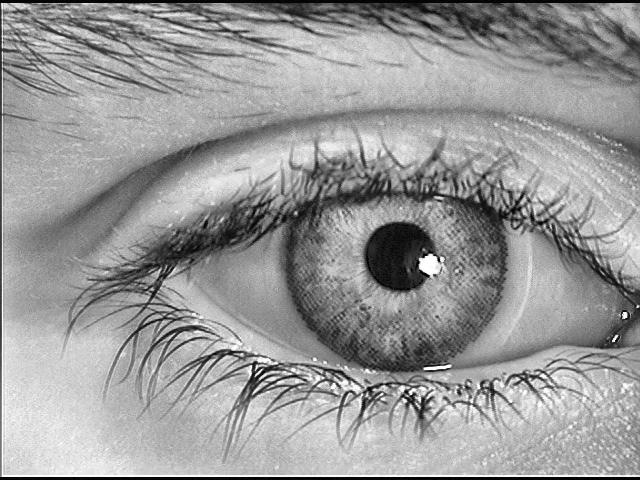}     
    \end{tabular}
    \caption{Clarkson Dataset: In the first and second columns we present real and fake images, respectively.}
    \label{fig:clarkson}
\end{figure}

The training set comprises a total of 4937 images: 2469 live images from 25 subjects, 1346 printed images from 13 subjects as well as 1122 patterned contact lens images from 5 subjects. The testing set includes additional unknown data and consists of 4066 images: 908 printed images, 765 patterned contact images, and 1485 live iris images.
In Figure~\ref{fig:clarkson}, we present some examples of live and fake images.

\subsection{Warsaw Dataset}

The Warsaw dataset consists of authentic iris images acquired by the IrisGuard AD 100 sensor and their corresponding paper printouts. The training set includes 4513 images: 1844 live images from 322 distinct eyes and 2669 images of the corresponding paper printouts. The testing set is composed of two subsets: known and unknown spoof images. 

\begin{figure}
    \centering
    \begin{tabular}{@{}cc@{}}
    \includegraphics[width=.47\linewidth]{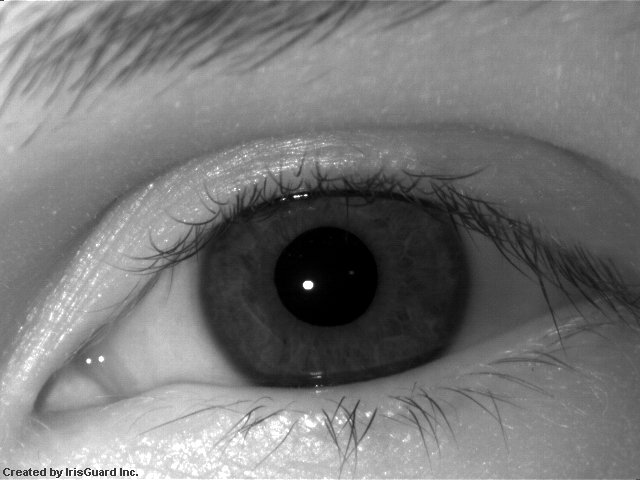} &
    \includegraphics[width=.47\linewidth]{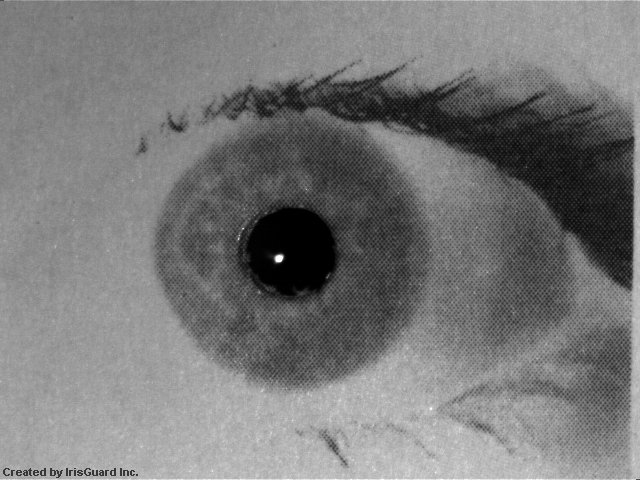} \\
    \includegraphics[width=.47\linewidth]{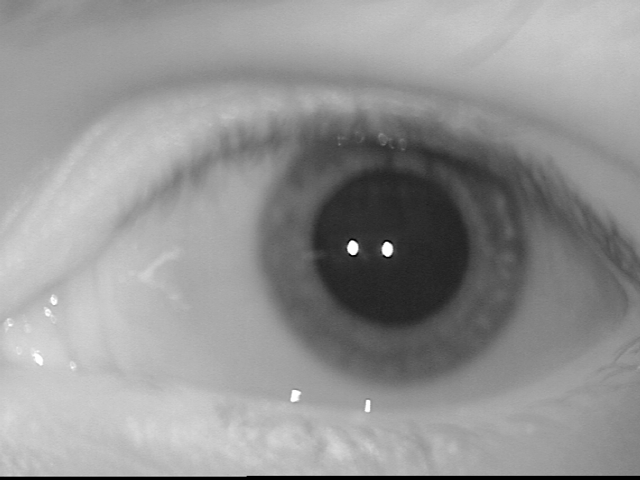} &
    \includegraphics[width=.47\linewidth]{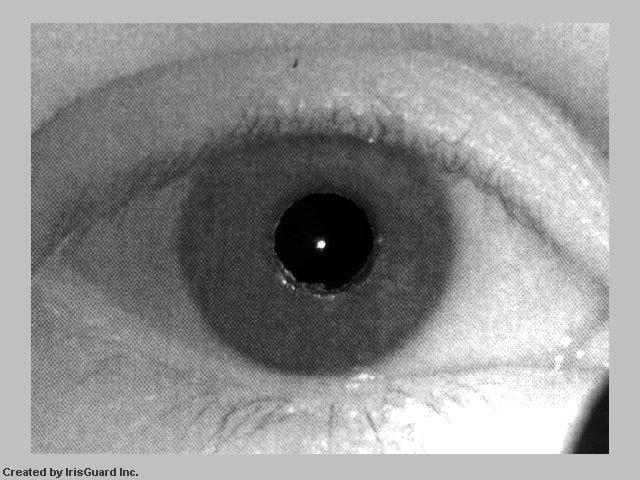} \\
    \includegraphics[width=.47\linewidth]{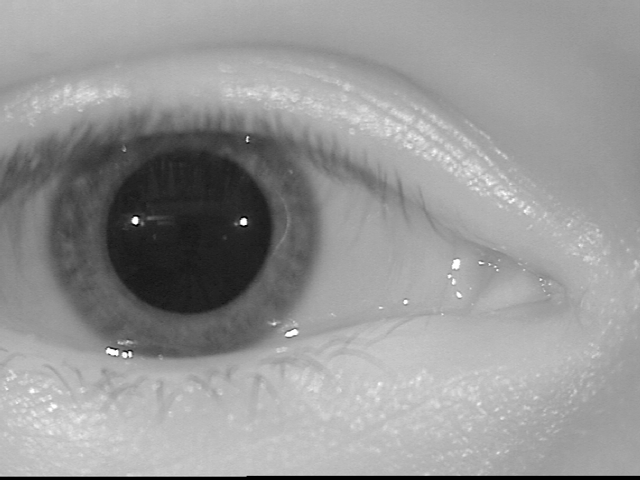} &
    \includegraphics[width=.47\linewidth]{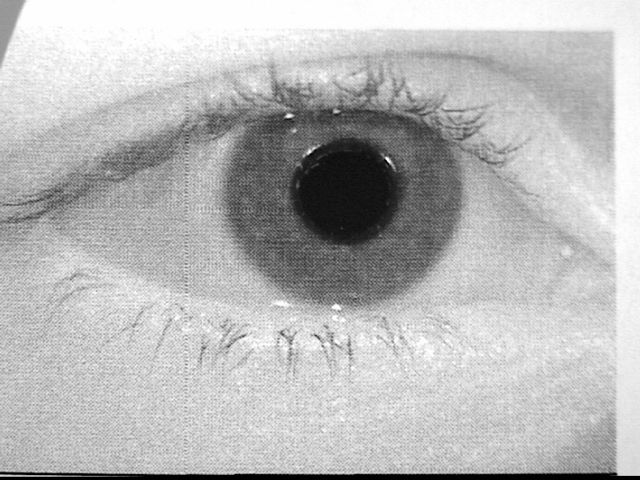}     
    \end{tabular}
    \caption{Warsaw Dataset: In the first and second rows we present real and fake images, respectively.}
    \label{fig:warsaw}
\end{figure}

Known spoofs subset includes 974 live iris images acquired from 50 distinct eyes and 2016 images of the corresponding printouts. The Unknown spoofs subset includes 2350 live iris images acquired from 98 distinct eyes and 2160 images of the corresponding printouts.
In Figure~\ref{fig:warsaw}, we present some examples of live and fake images.

\subsection{IIITD-VWU Dataset}

The IIITD-WVU dataset consists of samples acquired using the IriShield MK2120U mobile iris sensor at two different acquisition environments: indoors (controlled illumination) and outdoors (varying environmental situations). 

\begin{figure}[!tb]
    \centering
    \begin{tabular}{@{}cc@{}}
    \includegraphics[width=.47\linewidth]{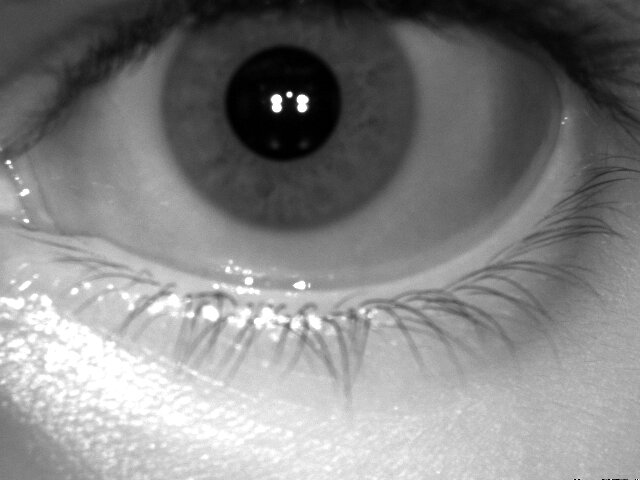} &
    \includegraphics[width=.47\linewidth]{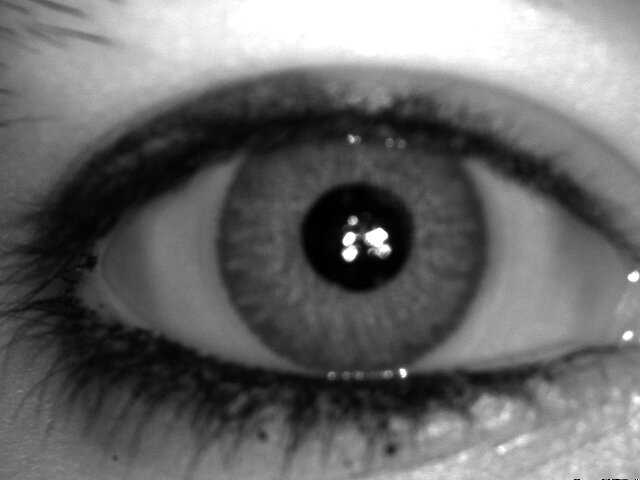} \\
    \includegraphics[width=.47\linewidth]{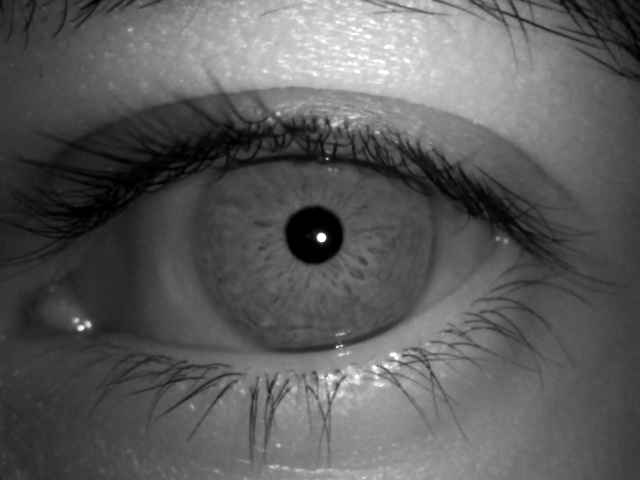} &   % real2
    \includegraphics[width=.47\linewidth]{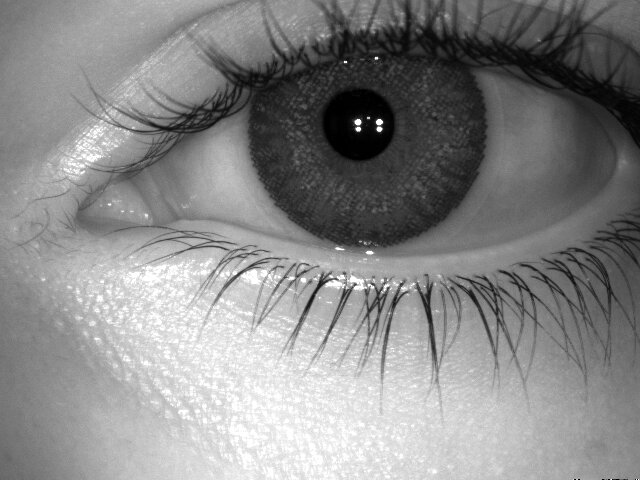} \\
    \includegraphics[width=.47\linewidth]{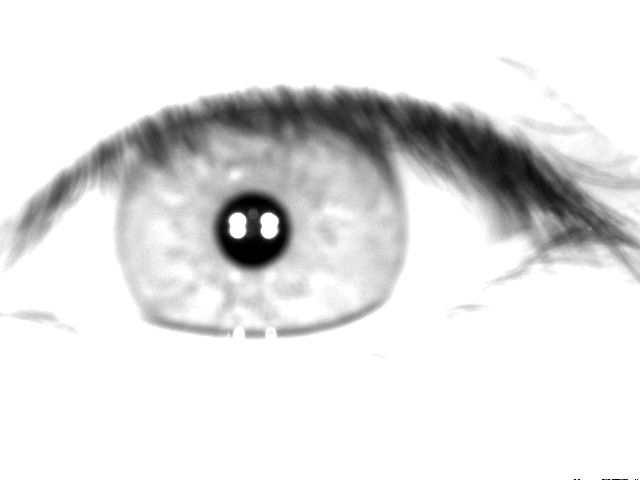} &   % real3
    \includegraphics[width=.47\linewidth]{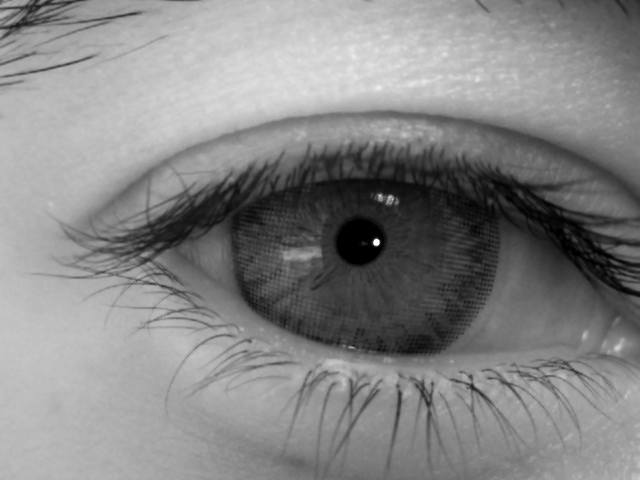}     % fake3
    \end{tabular}
    \caption{IIITD-WVU Dataset: In the first and second rows we present real and fake images, respectively.}
    \label{fig:iiitd-wvu}
\end{figure}

The training subset comprises 6250 images, combining  2,250 images of authentic irises and 4000 attack iris images, including textured contact lens iris images, printouts of live iris images, and printouts of contact lens iris images. Besides, the testing subset includes 4,209 iris images, 702 live iris images and 3507 attack iris images, combining textured contact lens iris images, printouts of live iris images and printouts of contact lens iris images. 
In Figure~\ref{fig:iiitd-wvu}, we present some examples of live and fake images.

\begin{figure*}[!tb]
\centering
\includegraphics[width=\linewidth]{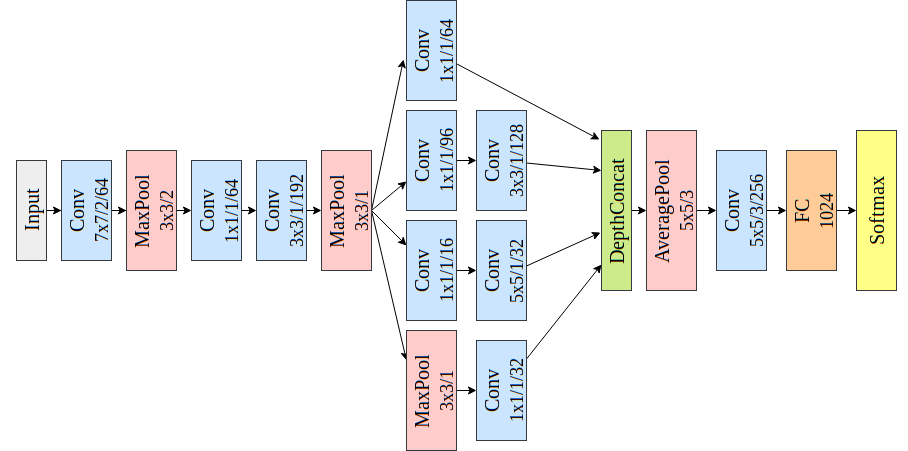}
\vspace{12pt}
\caption{CASIA SpoofNet architecture.}
\label{fig:SpoofNet}
\end{figure*}

\subsection{Notre Dame Dataset}

The Notre Dame dataset consists of samples acquired by LG 4000 and AD 100 sensors. The training subset consists of 600 images of authentic irises and 600 images of textured contact lenses manufactured by Ciba, UCL, and ClearLab. 
The  testing subset is split into known spoofs and unknown spoofs. 

\begin{figure}[!tb]
    \centering
    \begin{tabular}{@{}cc@{}}
    \includegraphics[width=.47\linewidth]{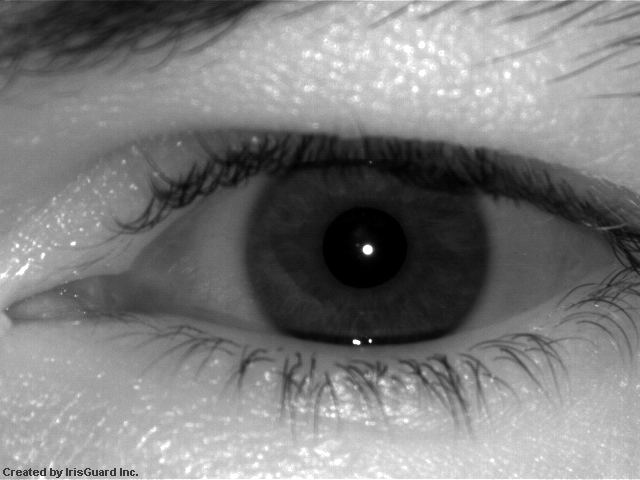} &
    \includegraphics[width=.47\linewidth]{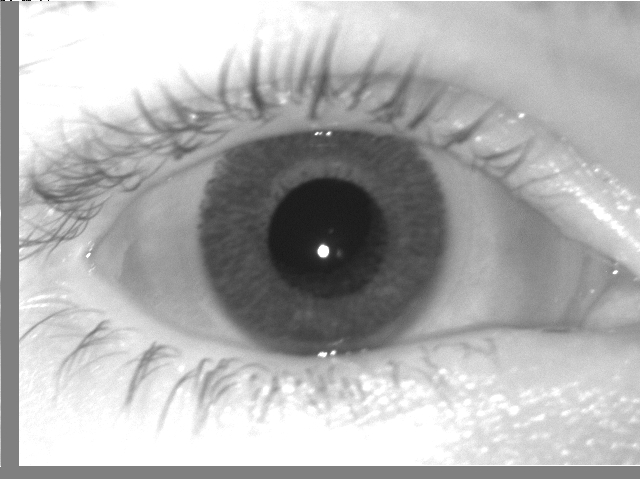} \\
    \includegraphics[width=.47\linewidth]{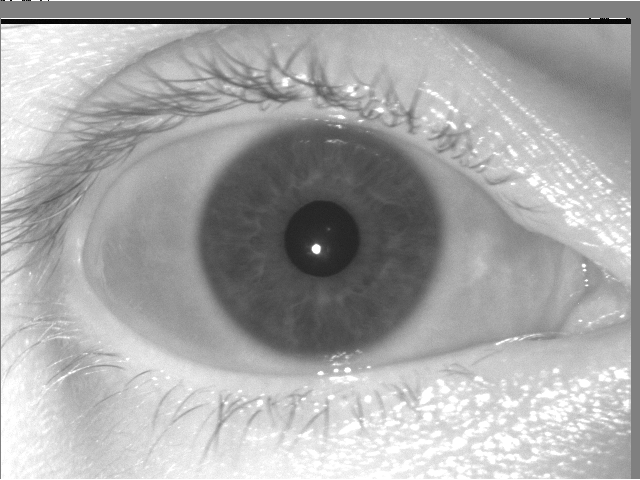} &
    \includegraphics[width=.47\linewidth]{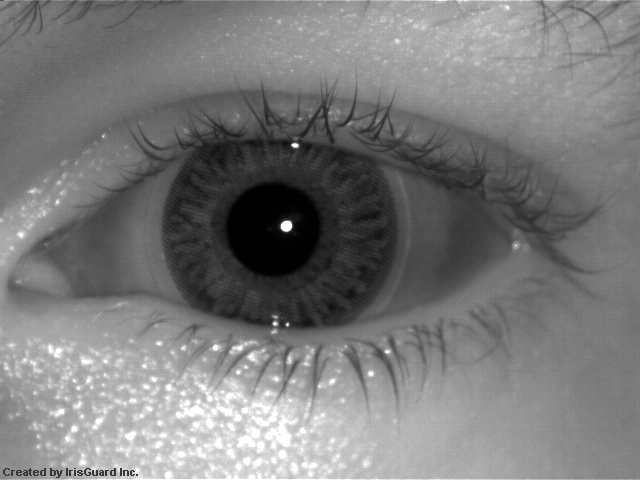} \\
    \includegraphics[width=.47\linewidth]{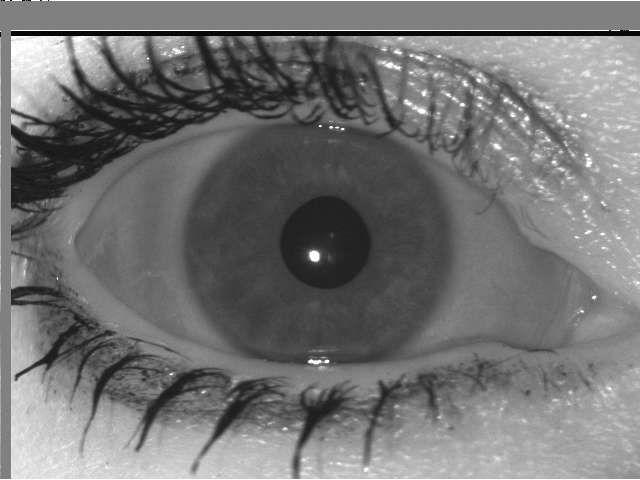} &
    \includegraphics[width=.47\linewidth]{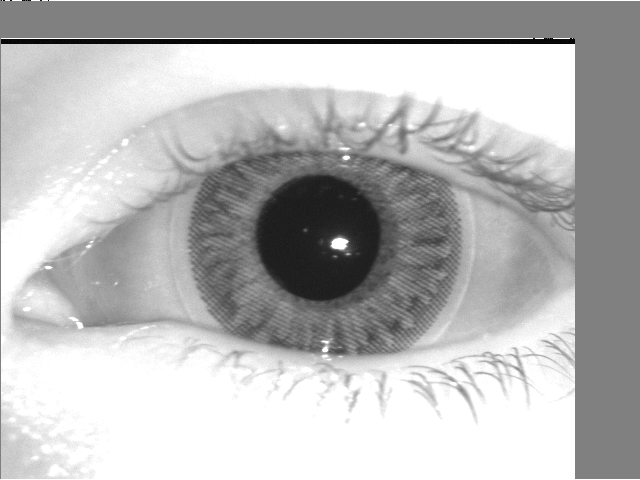}     
    \end{tabular}
    \caption{Notre Dame Dataset: In the first and second rows we present real and fake images, respectively.}
\vspace{12pt}    
    \label{fig:notredame}
\end{figure}

The known spoofs subset includes 900 images of textured contact lenses and 900 images of authentic irises. The unknown spoofs subset includes 900 images of textured contact lenses and 900 images of authentic irises.
In Figure~\ref{fig:notredame}, we present some examples of live and fake images.

\section{\uppercase{Proposed Algorithm}}

The CASIA algorithm submitted to the competition considers two types of iris spoofings: a printed iris and an iris with printed contact lenses. Since it is difficult to identify printed iris of poor quality, the printed iris and iris with contact lens are individually categorized. To this end, CASIA algorithm designed a Cascade SpoofNet combining two convolutional neural networks in sequence, SpoofNet-1 and  SpoofNet-2. SpoofNet-1 aims to distinguish printed and non-printed iris images. If the iris image inputted into the net is classified as a live sample, the iris is located and the re-scaled iris image is classified by the SpoofNet-2, to whether the sample is a live iris or a contact lens.

\begin{table*}[!tb]
\centering
\vspace{12pt}
\caption{Competition comparison - Error rates by dataset.}
\resizebox{0.98\textwidth}{!}{
\begin{tabular}{|c|c|cc|cc|cc|cc|cc|}
\hline
\multirow{2}{*}{Algorithm} & \multirow{2}{*}{Threshold} & \multicolumn{2}{c|}{Clarkson} & \multicolumn{2}{c|}{Warsaw} & \multicolumn{2}{c|}{IIITD-WVU} &
\multicolumn{2}{c|}{Notre Dame} & \multicolumn{2}{c|}{Combined} \\
 &  & APCER & BPCER & APCER & BPCER & APCER & BPCER & APCER & BPCER & APCER & BPCER \\ \hline
CASIA & 50 & 9.61\% & 5.65\% & 3.40\% & 8.60\% & 23.16\% & 16.10\% & 11.30\% & 7.56\% & 11.88\% & 9.47\% \\ \hline
\multirow{5}{*}{Method Proposed} 
 & 30 & 5.49\% & 0.00\% & 0.43\% & 2.94\% & 0.62\% & 31.9\% & 23.33\% & 0.27\% & 9.63\% & 1.49\% \\
 & 40 & 4.96\% & 0.00\% & 0.23\% & 3.82\% & 0.51\% & 34.61\% & 20.44\% & 0.72\% & 8.71\% & 1.90\% \\
 & 50 & 4.18\% & 0.00\% & 0.14\% & 4.60\% & 0.34\% & 36.89\% & 18.05\% & 0.94\% & 7.93\% & 2.32\% \\
 & 70 & 3.16\% & 0.06\% & 0.02\% & 6.70\% & 0.11\% & 41.31\% & 13.30\% & 1.40\% & 6.56\% & 3.37\% \\
 & 80 & 2.57\% & 0.20\% & 0.02\% & 8.40\% & 0.02\% & 43.58\% & 10.20\% & 2.20\% & 5.74\% & 4.15\% \\
 & 90 & 1.79\% & 0.47\% & 0.00\% & 9.05\% & 0.0\% & 49.85\% &    6.83\% &  3.20\% & 4.80\% & 5.64\% \\ \hline
\end{tabular}
}
\label{table:scores}
\end{table*}

\begin{table*}[!tb]
\centering
\caption{Cross-dataset experiment - Error rates by dataset.}
\resizebox{0.98\textwidth}{!}{
\begin{tabular}{|c|c|cc|cc|cc|cc|}
\hline
\multirow{2}{*}{Algorithm} & \multirow{2}{*}{Threshold} & \multicolumn{2}{c|}{Clarkson} & \multicolumn{2}{c|}{Warsaw} & \multicolumn{2}{c|}{IIITD-WVU} &
\multicolumn{2}{c|}{Notre Dame}  \\
 &  & APCER & BPCER & APCER & BPCER & APCER & BPCER & APCER & BPCER \\ 
 \hline
\multirow{5}{*}{Method Proposed} 
 & 30 & 37.0\% & 0.0\% & 0.0\% & 66.6\% & 41.3\% & 0.7\% & 30.2\% & 1.6\% \\
 & 40 & 35.2\% & 0.0\% & 0.0\% & 67.9\% & 37.2\% & 0.9\% & 27.7\% & 2.4\% \\
 & 50 & 33.0\% & 0.0\% & 0.0\% & 70.0\% & 34.1\% & 1.1\% & 25.8\% & 2.6\% \\
 & 70 & 28.6\% & 0.0\% & 0.0\% & 75.0\% & 26.9\% & 2.3\% & 20.9\% & 3.4\% \\
 & 80 & 25.8\% & 0.1\% & 0.0\% & 77.8\% & 22.4\% & 3.1\% & 18.2\% & 3.6\% \\
 & 90 & 22.5\% & 0.1\% & 0.1\% & 81.4\% & 16.1\% & 4.8\% & 14.6\% & 4.6\% \\
\hline
\end{tabular}
}
\label{table:cross}
\end{table*}

Both SpoofNets are based on GoogleNet \cite{r10} and consist of four convolutional layers and one inception module. The inception module is composed by layers of convolutional filters of dimension $1 \times 1$, $3 \times 3$ and $5 \times 5$, executed in parallel. It has the advantage of reducing the complexity and improving the efficiency of the architecture,  once the filters of dimension 1x1 help reduce the number of features before executing layers of convolution with filters of higher dimensions. The architecture proposed and their parameters (patch size/stride/filter dimension) is presented in Figure~\ref{fig:SpoofNet}. 

In this paper, the iris of the images was located manually in bounding boxes, using the graphical image annotation tool \textit{LabelImg} (this process could be done automatically but not so accurately using existing iris detection approaches~\cite{severo2018benchmark,lucio2019simultaneous}). 
The model was trained by 80\% of the training images and their corresponding iris images. The rest 20\% were used as the validation subset. In addition, the model was parameterized based on the analysis of empirical experiments.

During training, many hyperparameters can influence the effectiveness of the resulted model. Too many epochs might overfit the training data, while too few may not give enough time for the network to learn good parameters. 
To monitor the model's performance, the method \textit{EarlyStopping} from the \textit{Keras API} was used with the loss function binary cross-entropy. The patience parameter was set to 5 and training of the model was executed with a maximum of 20 epochs and batch size of 8.

The learning rate hyperparameter controls the speed at which the model learns. Specifically, it controls the amount of allocated error that the weights of the model are updated. The lower the value, the slower the model learns a problem. 
Typical values of the learning rate are less than 1 and greater than $10\mathrm{e}^{-6}$ \cite{r11}. 
In this work, training was better with a learning rate of value $10\mathrm{e}^{-5}$.

Finally, to reduce overfitting and improve the performance of the model on new data, the default value of weight decay was changed to $10^{-4}$ and a dropout of 0.2 was employed for regularization. 
%due to time constraints
The main optimal parameters are summarized in Table \ref{table:hyperparameters}. 
\begin{table}[!htb]
\centering
\caption{SpoofNet Hyperparameters Tuning}
\begin{tabular}{|c|c|}
\hline
Parameter              & Values \\ \hline
Number of epochs (max) & 20     \\ \hline
Batch size             & 8      \\ \hline
Learning rate          & $1\mathrm{e}^{-5}$   \\ \hline
Weight decay           & $10^{-4}$   \\ \hline
\end{tabular}
\label{table:hyperparameters}
\end{table}

\section{\uppercase{Performance Evaluation}}

The performance of the algorithm was evaluated based on the APCER, i.e, the rate of misclassified attack images, and the BPCER, i.e., the rate of misclassified live images. 
Both APCER and BPCER were calculated for each dataset individually, as well as for the combination of all datasets. 

Table~\ref{table:scores} shows the error rates obtained by both CASIA algorithm and the proposed modified CASIA algorithm, in the evaluation of the datasets individually and combined. 

In the competition, the threshold value for determining liveness was set at 50.
Comparing the error rates between the two algorithms, the proposed method performed better overall, with the only exception in the evaluation of the Notre Dame dataset, where the proposed method presented a higher APCER.
Besides that, Clarkson dataset showcased the lowest error rates, with a combined error rate of 4.18\% APCER and 0\% BPCER.

The presence of unknown spoofing attacks in the testing subsets was the main difficulty during the process of evaluation. 
Examining Table~\ref{table:scores} and analyzing the error rates presented by the proposed method, it is noticeable the significant trade-off between APCER and BPCER, once the model is not able to accurately classify real and unknown attack variations.

Aiming to mitigate this trade-off, different values of threshold were evaluated. 
Table~\ref{table:scores} shows the error rates presented by the proposed model to the thresholds of values 30, 40, 70, 80 and 90. 

The thresholds of values 70 to 90 helped reduce gradually the difference of the error rates by reducing the APCER and increasing the BPCER of the datasets of Notre Dame and  Clarkson university and for all the datasets combined, but did not have the same effect on the Warsaw dataset, where it caused the increase of the BPCER and the trade-off between the error rates. In this dataset, however, the thresholds of values 30 and 40 had a better effect, increasing the APCER and reducing both the value of BPCER and the trade-off with the dataset's APCER.
Note that for the IIITD-WVU dataset the results obtained were the worst among the compared datasets regarding the BPCER

We also performed a cross-dataset experiment, that is we trained the model using the data/images of all datasets except the one where it is evaluated/tested.
The goal here is to observe the generalization power of deep learning models and also verify how useful these models are for realistic scenarios where there is no data for a fine-tuning process.  
The results of this experiment are summarized in Table~\ref{table:cross}.
As can be observe by comparing the figures of Table~\ref{table:cross} and Table~\ref{table:scores}, the models learned using this cross-dataset scheme performed worse than the ones learned on its own data.
The less loss of performance is observed on the Notre Dame dataset, anyway the figures of APCER metrics are not suitable for some real-world applications.

\section{\uppercase{Conclusion}}

In this paper, we proposed a hyperparameter tuning of the neural network presented by the CASIA algorithm, submitted to the Iris Liveness Detection competition of 2017. 
Most of the databases evaluated included unknown presentation attacks, being the main difficulty to our model. 
The suggested modifications significantly reduced the values of APCER and BPCER of the datasets. 

The Clarkson dataset showcased the lowest error rate, with 4.18\% APCER and 0\% BPCER for the threshold of value 50. 
In an attempt to reduce the trade-off between the values of APCER and BPCER, different threshold values were evaluated. The thresholds of values 70 to 90 worked out successfully for the datasets of Clarkson and Notre Dame University and all the datasets combined, reducing gradually the dataset's APCER. 
On the other hand, the thresholds of values 30 and 40 had a better effect on the Warsaw dataset, reducing the BPCER and the trade-off with the dataset's APCER.

\section*{\uppercase{Acknowledgments}}

The authors would like to thank 
the Brazilian National Research Council -- CNPq (Grants \#313423/2017-2  and \#428333/2016-8); 
the Foundation for Research of the State of Paraná (\textit{Fundação Araucária})
the Coordination for the Improvement of Higher Education Personnel (CAPES) (Social Demand Program);
and also acknowledge the support of NVIDIA Corporation with the donation of the Titan Xp GPU used for this research.

\balance
\bibliographystyle{apalike}
{\small
\bibliography{main}}

\end{document}